\documentclass{article}

\usepackage{natbib}

\usepackage[preprint]{nips_2018}

\usepackage[utf8]{inputenc} % allow utf-8 input
\usepackage[T1]{fontenc}    % use 8-bit T1 fonts
\usepackage{hyperref}       % hyperlinks
\usepackage{url}            % simple URL typesetting
\usepackage{booktabs}       % professional-quality tables
\usepackage{amsfonts}       % blackboard math symbols
\usepackage{nicefrac}       % compact symbols for 1/2, etc.
\usepackage{microtype}      % microtypography
\usepackage{caption}
\usepackage{graphicx, subfig}
\usepackage{amsmath,amsthm,amssymb}
\usepackage{multirow}

\def\bk{{\bf k}}
\def\bq{{\bf q}}
\def\bs{{\bf s}}
\def\RB{{\mathbb R}}

\def\softmax{\mathrm{softmax}}
\def\Attention{\mathrm{Attention}}
\def\MultiHead{\mathrm{MultiHead}}
\def\head{\mathrm{head}}
\def\Concat{\mathsf{Concat}}
\def\Ham{\mathrm{Ham}}
\def\Bleu{\mathrm{BLEU}}

\def\maxl{\mathop{\max}\limits}
\def\minl{\mathop{\min}\limits}

\title{Hierarchical Attention: What Really Counts in Various NLP Tasks}

\author{
  Zehao Dou\\
  School of Mathematical Science\\
  Peking University\\
  Beijing, 100871 \\
  \texttt{zehaodou@pku.edu.cn} \\
  \And
  Zhihua Zhang\\
  School of Mathematical Science\\
  Deep Learning Lab\\
  Peking University\\
  Beijing, 100871
  \texttt{zhzhang@math.pku.edu.cn} \\
}

\begin{document}

\maketitle

\newtheorem{Theorem}{Theorem}

\begin{abstract}
Attention mechanisms in sequence to sequence models  have shown great ability and wonderful performance in various natural language processing  (NLP)  tasks, such as sentence embedding, text generation, machine translation, machine reading comprehension, etc. Unfortunately, existing attention mechanisms only learn either high-level or low-level features. In this paper, we think that the lack of hierarchical mechanisms is a bottleneck in improving the performance of the attention mechanisms, and propose a novel \textbf{Hierarchical Attention Mechanism (Ham)} based on the weighted sum of different layers of a multi-level attention. 
\textbf{Ham} achieves a state-of-the-art BLEU score of $0.26$ on Chinese poem generation task and a nearly $6.5\%$ averaged improvement compared with the existing machine reading comprehension models such as BIDAF and Match-LSTM. Furthermore, our experiments and theorems reveal that  \textbf{Ham} has greater generalization and representation ability than existing attention mechanisms. 

%Applying this Hierarchical Attention mechanism to several famous models of various NLP tasks, 
%%including Chinese poem generation, machine reading comprehension and machine translation, 
%we achieve a new state-of-the-art BLEU score in Chinese poem generation task, and a nearly $6.5\%$ averaged improvement compared with the existing machine reading comprehension models such as BIDAF and Match-LSTM. From all our experiments and theorems, we can say that the Hierarchical Attention mechanism has greater generalization and expressing ability than existing attention mechanisms. 
\end{abstract}

\section{Introduction}

In recent years, long short-term memory, convolutional and recurrent networks with attention mechanisms have been 
very successful on a variety of NLP tasks.
%applied to various tasks of NLP and showed excellent or even state-of-the-art performance. 
Most of the tasks In NLP can be formulated as sequence to sequence problems.  Moreover, encoders and decoders are vital components in sequence to sequence models while processing the inputs and outputs. 

An Attention mechanism works as a connector between the encoders and decoders and
help the decoder to decide which parts of the source text to pay attention to. 
%model to pay attention to important parts of a text and to overlook less important parts. 
%In other words, attention mechanism can
Thus an attention mechanism can integrates sequence models and transduction models, so it is able to connect two words in a single passage or paragraph without regard to their positions. 

Attention mechanisms have become integral  components in various NLP models.    
For example, in machine translation tasks, attention mechanism-based models  [1, 2, 3] have ever been the state-of-the-art; in sentence embedding, self attention based model is now the state-of-the-art [4]; in machine reading comprehension, almost every recently-published model, such as BIDAF[5], Match-LSTM[6], Reinforcement Ranker Reader[7], and R-NET[8], contains attention mechanism; in abstractive summarization model [12] which has also once been the state-of-the-art, attention mechanism is very necessary, and in poem generation [9], attention mechanism is also widely used. More surprisingly, Vaswani, et al.\ (2017) showed  that their model \textbf{Transformer} which relies solely on the attention mechanisms can outperform those RNN or LSTM-based existing models in machine translation tasks. Thus, they stated that "Attention is all you need". 

However, we note that the potential issue with the existing attention mechanisms is that the basic attention mechanism learns only the low-level features while the multi-level attention mechanism learns only the high-level features. This may make it difficult for the model to capture the intermediate feature information, especially when the source texts are long. In order to address this issue, we present \textbf{Ham} which introduces a hierarchical mechanism into the existing multi-level attention mechanisms. Each time when we perform a multi-level attention, instead of using the result of the last attention level only, we use the weighted sum of the results of all the attention levels as the final output.
%So far as we  know, there are no hierarchical mechanisms. 

%That is why we want to create a hierarchical attention model to learn all levels of features among the tokens in the input sequence, which can make our encoder output a more suitable and learnable intermediate results for decoder.
We show that \textbf{Ham} can learn all levels of features among the tokens in the input sequence and give a proof of its monotonicity and convergency.  
This work presents the design and implementation of  \textbf{Ham}  and our implementation performs well on a range of tasks by replacing the existing attention mechanisms in different models of different tasks. 

We are able to achieve results comparable to or better than existing state-of-the-art model. On Chinese poem generation, our model scores $0.246$ BLUE, an improvement of $21.78\%$ from a RNN-based Poem Generator model.
On machine reading comprehension task, our implementation is more effective, model with \textbf{Ham} has achieved an average improvement of $6.5\%$ compared to previous models.

%In this paper we introduce our novel Hierarchical Attention Mechanism and give a proof of its monotonicity and convergency. Moreover, by replacing the existing attention mechanisms in different models of different tasks, we show the powerfulness of our new mechanism. In Chinese poem generation, we achieve a new state-of-the-art BLEU score; and in machine reading comprehension task, we get a nearly $6.5\%$ averaged improvement in various models after replacing Attention Mechanism with \textbf{Hierarchical Attention Mechanism}.

The implementation of the Hierarchical Attention Mechanism is not difficult and the code will be available on \url{http://github.com} after the acceptance.

\section{Attention mechanisms}
%\subsection{Attention}

The attention mechanism can be described as a function whose input is a query and a set of key-value pairs, where the query and keys are vectors with the same dimension (denoted $d_{k}$), and  the values are defined  as $d_{v}$-vectors.  Note that in most types of attention mechanisms, the values are equal to the keys. Through the mapping of the attention mechanism, the input can be mapped to a single vector, which is as the output. 

\subsection{The Vanilla Attention Mechanism(VAM)}

Given a query $\bq \in \RB^{d_k}$ and an input sequence $K = [\bk_{1}, \bk_{2},  \ldots , \bk_{n}]\in \RB^{d_{k}\times n}$ where $\bk_{i}\in \RB^{d_{k}}$ denotes the word embedding of the $i$-th word of the sequence,  the vanilla attention mechanism aims at using a compatibility function $f({\bk_{i}}, \bq)$ to compute a relativity score between the query $\bq$ and each word $\bk_{i}$.  This score is treated as the attention value of $\bq$ to $\bk_{i}$. Then we have $n$ attention scores $f(\bk_{i}, \bq)$ for $i = 1,2, \ldots, n$. Now we apply the softmax function to define a categorical distribution: 
\[
p(i | K,  \bq) = \softmax(f(\bk_{i}, \bq)) = \frac{\exp (f(\bk_{i}, \bq))}{\sum_{j=1}^{n} \exp(f(\bk_{j}, \bq))}.
\]
Futher, we compute the output which  is represented as the weighted sum of the input sequence:
\[\bs = \sum_{i = 1}^{n}p(i | K,  \bq) \bk_{i}.
\]
The attention mechanism above is the original version which was firstly proposed by Bahdanau, et al.\ (2014). In the \textbf{Scaled Dot-product Attention Mechanism},  the compatibility function $f$ is defined as the scaled dot product function $f(\bk_{i}, \bq) = \frac{<\bk_{i}, \bq>}{\sqrt{d_{k}}}$. Here the scaling factor $\frac{1}{\sqrt{d_{k}}}$ is used to prevent the dot product from growing too large in magnitude.  

\subsection{Soft, Hard and Local Attention Mechanisms}

There are  three different types of mechanisms:  soft, hard and local. The main difference between them is the region where attention function is calculated. The \textbf{VAM}  belongs to  soft attention where the categorical distribution $p(i|{K}, \bq)$ is computed over the whole input sequence of words. Thus it is also referred to as global attention. The resulting distribution can reflect the relatedness or importance between the query and every word in the input sequence, and we use these importance scores as the weights of the output sum. Soft attention takes every word in the input sequence, no matter what kind of word it is, into consideration.  Soft attention is  differentiable and  parameter-free, but 
is computationally expensive and less accurate.

In order to overcome the weakness of soft attention,  hard attention is a natural alternative.  Contrary to the widely-studied soft attention, hard attention locates accurately to only one key $k_{i_{0}}$. In other words, the probability of getting the special key $k_{i_{0}}$ is 1 and others be 0. This implies that the choice of the one key means everything to the performance of the model. The action of choosing is not differentiable, so one uses reinforcement learning methods instead, such as policy gradient method.

As we have seen, soft and hard attentions  are two extreme cases. Xu, et al.\ (2015) proposed a hybrid attention mechanism. Instead of choosing every key or only one key, one chooses a subset of all the keys from the input sequence. When computing the attention, one can just focus on the important part of the keys and discard the rest, thus it is also referred to as local attention. This attention mechanism combines the wideness and  accuracy when choosing keys. The subset-choosing process is non-differentiable and reinforcement learning methods are also needed.

\subsection{Multi-Head Attention and Multi-Level Attention Mechanisms}

The multi-head attention mechanism proposed by Vaswani, et al.\ (2017)  plays an important role in the \textbf{Transformer} model which is state-of-the-art in neural machine translation. Instead of calculating a single attention function with queries, keys and values, it linearly projects the queries, keys and values $h$ times  to $d_{k}$, $d_{k}$ and $d_{v}$ dimensions, respectively. On each version of linear projections, the attention function is performed in parallel and yields several versions of $d_{v}$-dimensional scaled dot-product attentions. Subsequently, these attention values are concatenated and once again projected, resulting in the final value as the output of the multi-head attention mechanism. That is, 
\[\Attention(Q,K,V) = \softmax(\frac{QK^{T}}{\sqrt{d_{k}}})V, \]
\[\MultiHead(Q,K,V) = \Concat(\head_{1}, \cdots, \head_{h})W^{O},
\]
where $\head_{i} = \Attention(QW_{i}^{Q}, KW_{i}^{K}, VW_{i}^{V})$ and $W_{i}^{Q},W_{i}^{K},W_{i}^{V},W_{i}^{O}$ are all projection matrices. \\

The multi-level attention mechanism is another variety of attention mechanisms. Instead of increasing the number of heads, the multi-level attention increases the number of levels. For example, in a two-level attention mechanism, we calculate the attention value of the query $\bq$ and the keys, which is represented as $\Attention(\bq, {K}, {K})$. This output has the same dimension as the query, giving the first level. In the second level, we treat the output as the new query and calculate the attention value with the input sequence (or keys) ${K}$ again. The result can be represented as $\Attention(\Attention(\bq, {K}, {K}),  {K}, {K})$. The second attention can learn a higher level of internal features among the words of the input text. 

Based on the long line of previous attempt, Cui et al.\ (2017) proposed a novel way of treating various documents in neural machine translation. They used a self-attention mechanism to encode the words in every document and then  used a second attention over different documents to learn a higher level of features among the words of different documents. Yang, et al.\ (2016) also proposed a 2-level hierarchical attention network for document classification task. In recently proposed \textbf{Transformer} mode [1],  the authors repeated the attention mechanism $N$ times over the input sequence in order to learn the higher level feature. This is an $N$-level attention mechanism through which the input sequences can be changed again and again into sequences much more suitable for feature extraction and decoder input. This is why so-called  \textbf{Transformer}. 

\subsection{Self Attention Mechanism}

In the self attention mechanism  the query and  key are the same. In other words,  the query $\bq$ stems from the input sequence ${K}$ itself. Using self attention mechanism, we are able to learn the relatedness of different parts of the input sequence, no matter what their distance is. With self-attention, a long text can be encoded to a more suitable input for the decoder. Similar to the original attention mechanism, self-attentions also have three different types: soft, hard and local, as well as  have multi-head version and multi-level version.

\section{Hierarchical Attention Mechanism\ (Ham)}

In this section we present two Hierarchical Attention models built on the vanilla attention and self attention, respectively.

\subsection{Hierarchical Vanilla Attention Mechanism (Ham-V)}

We have mentioned above that multi-level attention mechanisms can learn a deeper level of features among all the tokens of the input sequence and the query. In our model, we use multi-level for reference but different from the existing \textbf{Multi-Level Attention Mechanisms}.  Our \textbf{Ham-V} focus on all the intermediate attention results rather than just the result of the last attention level. As shown in Figure1, given the query and the input sequence which consists of $n$ keys, we calculate the \textbf{Vanilla Attention Mechanism} result of them and get Query 1. And then we continue to calculate the attention result of Query 1 and the keys and get Query 2. Repeat this calculation $d$ times. Thus,  we form a $d$-depth attention. Finally, the output of our \textbf{Ham-V} is the weighted sum of the above $d$ attention results, where the $d$ weights are the softmax values of $d$ trainable parameters. The softmax is used to convert these $d$ weights into  the probabilities. These weights can tell us the relative importance of the $d$ intermediate attention results Query $i$. In other words, the relative importance of the $d$ attention levels. 
\begin{figure}
\centering
\includegraphics[width=0.80\textwidth]{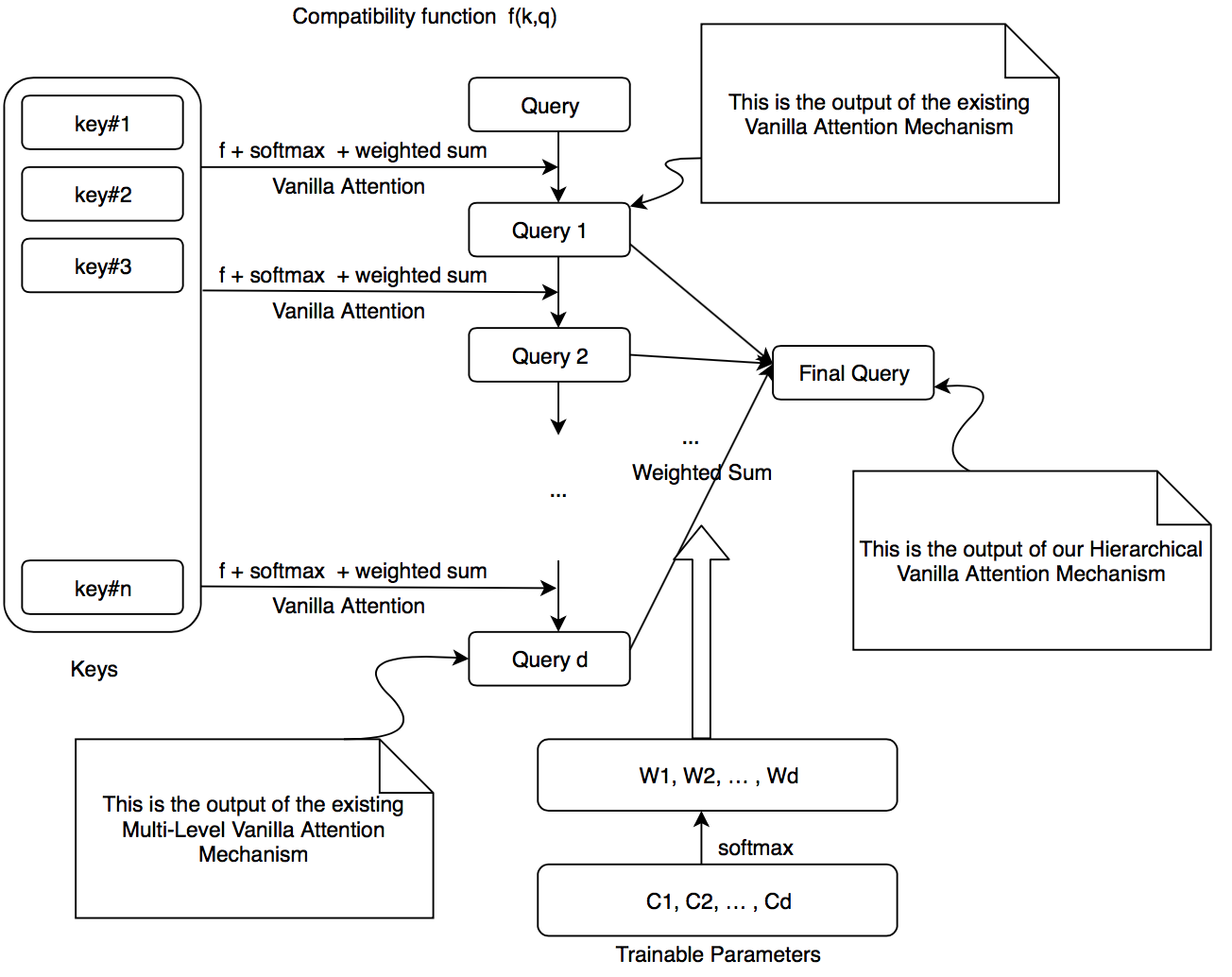}
\caption{Hierarchical Vanilla Attention Mechanism}
\label{img1}
\end{figure}

\subsection{Hierarchical Self Attention Mechanism (Ham-S)}

In \textbf{Self Attention Mechanisms}, the query stems from the input sequence ${\bk}$ itself. So it can be treated as a special case of attention mechanisms. Similarly, the self-version of \textbf{Hierarchical Attention Mechanisms}, which is shown in Figure 2, takes only the sequence ${\bk}$ as input. We calculate the self-attention results of the input sequence for $d$ times consecutively. %The integer number $d$ is called self-attention depth. 
Finally, the output of our \textbf{Ham-S} is the weighted sum of these $d$ attention results, where the $d$ weights are the softmax values of $d$ trainable parameters $w_{1},w_{2},\cdots, w_{d}$ which is the same as \textbf{Ham-V}. Through the $d$ levels of self-attention, our model can learn different levels of deep features among all the tokens of the input sequence, and through the $d$ trainable parameters and the weighted sum mechanism, our model can learn the relative importance of the $d$ self-attention levels. 

\begin{figure}
\centering
\includegraphics[width=0.80\textwidth]{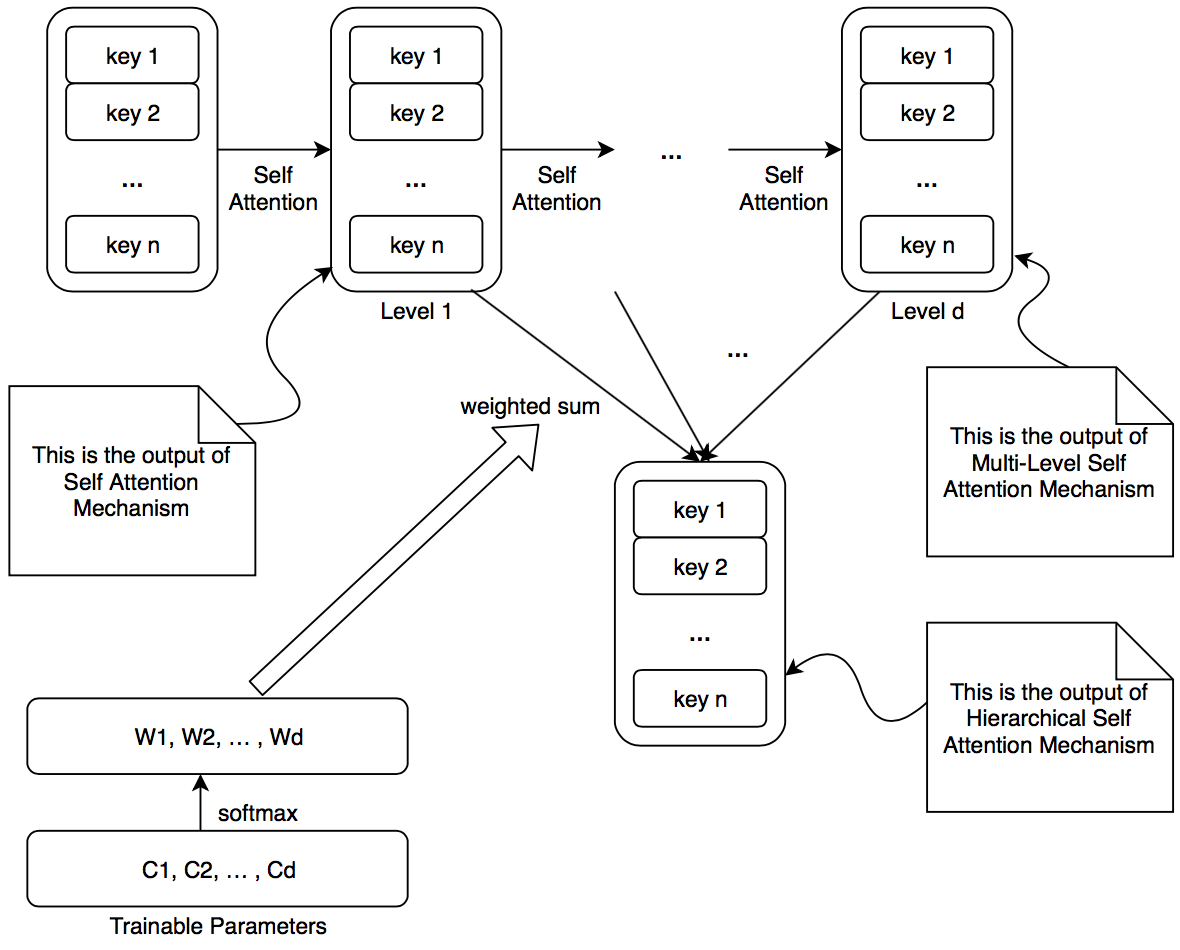}
\caption{Hierarchical Self Attention Mechanism}
\label{img1}
\end{figure}

\section{Analysis} %{of \textbf{Ham} Model}

%Before doing experiments, 

We present theoretical analysis  of our \textbf{Ham} mainly in two aspects: representation ability and convergence. 
The representation ability of \textbf{Ham} is obviously higher than \textbf{Vanilla Attention Mechanisms} and \textbf{Multi-Level Attention Mechanisms}. 
Because the two attention mechanisms are just two extreme cases of \textbf{Ham}.
%This is because 
When we set $w_{1}=1$, $w_{2}=w_{3}=\cdots=w_{d}=0$, our \textbf{Ham} model is equivalent to the former.  When we set  $w_{d}=1, w_{1}=w_{2}=\cdots=w_{d-1}=0$, our model is equivalent to the latter. Thus
\textbf{Ham} is much more general and 
 the representation ability  is much higher. Using the weighted linear combination of these $d$ intermediate attention results, our model takes every level of features into consideration.

We are going to prove that the global minimum value of our loss function $L(\Ham(c_{1},\cdots ,c_{d}), X, \theta)$ of the whole \textbf{Ham}  model  will decrease monotonically and converges finally as the increase of hierarchical attention depth $d$. Here, $\Ham(c_{1},\cdots ,c_{d})$ denotes a \textbf{Ham}  with the attention depth $d$ and  $d$ trainable parameters $c_{1}, \cdots ,c_{d}$,  $X$ denotes all the input data including the queries and keys,  and $\theta$ denotes all the parameters in the other part of the whole model. It is also worth emphasizing that all loss functions in NLP tasks have positive values.

\begin{Theorem}
In Vanilla Attention Mechanism, we have that 
\[ \minl_{1\leqslant i\leqslant n}\|\bk_{i}\|_{2} \leqslant \|\Attention(\bq, {K}, {K})\|_{2} \leqslant \maxl_{1\leqslant i\leqslant n}\|\bk_{i}\|_{2}.
\] %where ${K}=(\bk_{1}, \bk_{2},\cdots, \bk_{n})$.
\end{Theorem}
This result is very obvious
%This is a very obvious result because 
 according to the definition of \textbf{Vanilla Attention Mechanism}, $\Attention(\bq, {K}, {K})$ is the weighted sum of $\bk_{1}, \bk_{2},\cdots , \bk_{n}$ and the weights are nonnegative with their sum 1. Denote the weights as $\alpha_{1}, \cdots, \alpha_{n}$. Then
\[\|\Attention(\bq, {K}, {K})\|_{2}= \Big\|\sum_{i=1}^{n}\alpha_{i} \bk_{i}\Big\|_{2}\leqslant \sum_{i=1}^{n}\alpha_{i}\|\bk_{i} \|_{2} \leqslant \max_{1\leqslant i\leqslant n}\|\bk_{i}\|_{2}.
\]
The left-hand side of the inequality can be proved similarly. This theorem tells us that through multiple attention layers, the vectors of intermediate attention levels will neither explode nor vanish as the increase of attention depth $d$. 

\begin{Theorem}
Let $A_{d} = \minl_{c_{i}, \theta} L(\Ham(c_{1},\cdots ,c_{d}), X, \theta)$ be the global minimal value of the loss function. Then $\{A_{d}\}|_{d = 1}^{+\infty}$ is a monotonically decreasing sequence and it will converge.
\end{Theorem}
It is easy to note that:
%\begin{equation}
\begin{align*}
A_{d} &= \min_{c_{i}(1\leqslant i\leqslant d), \theta} L(\Ham(c_{1}, \cdots, c_{d}), X, \theta)\\
&=  \min_{c_{i}(1\leqslant i\leqslant d), \theta} L(\Ham (c_{1}, \cdots, c_{d}, c_{d+1}=-\infty), X, \theta)\\
&\geqslant \min_{c_{i}(1\leqslant i\leqslant d+1),\theta}L(\Ham(c_{1}, \cdots, c_{d}, c_{d+1}), X, \theta) = A_{d+1}.
\end{align*}
%\end{equation}
This means the monotonicity of the sequence $\{A_{d}\}|_{d = 1}^{+\infty}$. On the other hand, since our loss function always has positive values, sequence $\{A_{d}\}$ has 0 as its lower bound. Therefore, the monotonically decreasing sequence $\{A_{d}\}|_{d = 1}^{+\infty}$ will converge.

\section{Experiments}

Attention mechanisms are widely used in various NLP tasks. In our experiments, we would like to replace existing attention mechanisms with our novel \textbf{Ham-V} and replace existing self-attention mechanisms with our \textbf{Ham-S} to show the powerfulness and generalization ability of our model. We conduct our experiments on two different NLP tasks, Chinese Poem Generation and Machine Reading Comprehension. 
%and Neural Machine Translation, which cover the most popular and difficult missions. 

\subsection{Machine Reading Comprehension}

The first NLP task we used to test our \textbf{Ham} is the machine reading comprehension (MRC). We conduct our experiment on both English MRC and Chinese MRC. The baseline models we use include BIDAF[5], Match-LSTM[6], R-NET[8]. Here, BIDAF is for both Chinese and the other two are for English. They have a major similarity that  all of them use attention mechanism as the connection between their encoders and decoders and they are all open source. Their code is available on \url{http://github.com/baidu/DuReader}, \url{https://github.com/MurtyShikhar/Question-Answering} and \url{https://github.com/NLPLearn/R-net}. What we will do is to replace their attention mechanism with \textbf{Ham} and compare the difference of their performance. Specially, the R-NET model contains two attention mechanisms when doing question-passage matching and passage self-matching. Here, we replace them both with \textbf{Ham} and \textbf{Ham-S}. 

The Chinese dataset we use for MRC experiments is DuREADER which is introduced by He, et al.\ (2017) and the English dataset we use include SQUAD which can be downloaded from \url{https://rajpurkar.github.io/SQuAD-explorer} and MS-MARCO which can be downloaded from \url{http://www.msmarco.org}. We randomly choose 10 percent of question-answer data as testing set and the rest as training set. The evaluation method we use is BLEU-4 and ROUGE-L for Chinese, ExactMatch(EM) and F1 score for English, where EM measures the percentage of how much the prediction of the model matches ground truth exactly and F1 measures the overlap between prediction and ground truth. During our experiments, we set also different attention depths $d$ to show the influence of attention depth.

\subsection{Chinese Poem Generation}

In this work we generate Chinese quatrains, each of whose lines has the same length of 5 or 7 characters. The baseline model we use is \textbf{Planning based Poetry Generation (PPG)} proposed by Wang, et al.\ (2016), which generates Chinese poetries with a planning based neural network. Once we input a Chinese text, this model will generate a highly-related Chinese quatrains as  follows. 

Firstly, the model extracts keywords from this input text with \textbf{TextRank algorithm} proposed by Mihalcea, et al.\ (2004). Next, if the number of extracted keywords is not enough for a whole quatrain, more keywords will be created by \textbf{Knowledge-based method}. Then it comes to the final step, poem generation. The quatrain is generated line by line and each line corresponds a keyword. When generating a single line, one uses a bi-directional Gated Recurrent Unit (GRU) model proposed by Cho, et al.\ (2014 ) as encoder and another GRU model as decoder. Between encoder and decoder, an attention mechanism is used for connection. 

It is worth emphasizing that  the dataset used by \textbf{PPG} consists of 76,859 quatrains from the Internet and \textbf{PPG} randomly chooses 2000 quatrains for testing, 2000 for validation and the rest for training. In the encoder part of \textbf{PPG}, the word embedding dimensionality is set as 512 and initialized by word2vec (Mikolov, et al.\ (2013)). In both GRU models, the hidden layers also contain 512 hidden units but they are initialized randomly. For more details, please read Wang, et al.\ (2016). The code and dataset of \textbf{PPG} model can be found from \url{https://github.com/Disiok/poetry-seq2seq}. \\

In our experiment, we replace the attention part of \textbf{PPG} from a Vanilla Attention Mechanism to our \textbf{Ham-V} and set the compatibility function $f$ to be the scaled dot product function, while keeping other parts and dataset unchanged as \textbf{PPG} except the evaluation part. The evaluation of poem generation in \textbf{PPG} is done by experts and we can not keep evaluation method unchanged since it is not convincing to find experts for evaluation. Our evaluation algorithm is based on BLEU-2 score which is calculated as
\[\Bleu = \frac{1}{3}\sum_{i=1}^{3} \Bleu_{i},
\]
where $\Bleu_{i}$ denotes the BLEU-2 score computed for the next $(i+1)$th lines given the previous $i$ goldstandard lines. This averaged $\Bleu$ can tell us how much correlated the lines of a generated quatrain are. 
We will show some quatrains generated by our \textbf{Ham-based PPG} in the appendix.

\section{Experimental Results and Qualitative Analysis}
\subsection{Machine Reading Comprehension}
As clearly visible in Table 1, the proposed model is much better than conventional model at Chinese and English machine reading comprehension. This is likely due to the fact that our human language has a kind of hierarchical relationship within itself both structurally and semantically. And our hierarchical attention mechanism is much easier to capture the inherent structural and semantical hierarchical relationship in the source texts because of their innate similarity.
%The results are summarized in Table 2. We compare our \textbf{Ham}-based MRC models with those ordinary \textbf{Ham}-free ones. 
We also set the attention depths $d$ to be 1 (which is equivalent to ordinary models without using \textbf{Ham}), 2, 5, 10 and 20.

From Table 1, we can find that our \textbf{Ham} plays a significant role of the whole model. With the increase of attention depth $d$, the performance rises quickly at first and starts to converge when $d$ grows larger. The biggest improvements on these three models are $7.26\%$, $7.76\%$ and $5.64\%$ respectively. Their average is over $6.5\%$ which is a huge progress. 
\begin{table}
\centering
\begin{tabular}{|l|c|c|c|}
\hline
Model & Dataset & BLEU-4 & ROUGE-L \\
\hline
BIDAF & DuReader & 33.95 & 44.20 \\
BIDAF with 2-level \textbf{Ham} & DuReader & 34.02 & 44.39\\
BIDAF with 5-level \textbf{Ham} & DuReader & 34.79 & 45.10\\
BIDAF with 10-level \textbf{Ham} & DuReader & \textbf{35.96} & 47.33\\
BIDAF with 20-level \textbf{Ham} & DuReader & 35.79 & \textbf{47.41}\\
\hline
\hline
Model & Dataset & EM & F1\\
\hline
Match-LSTM & SQUAD & 54.29 & 66.87 \\
Match-LSTM with 2-level \textbf{Ham}& SQUAD & 54.41 & 66.99\\
Match-LSTM with 5-level \textbf{Ham}& SQUAD & 55.29 & 69.03\\
Match-LSTM with 10-level \textbf{Ham}& SQUAD & 58.37 & 70.70\\
Match-LSTM with 20-level \textbf{Ham}& SQUAD & \textbf{58.47} & \textbf{70.81}\\
\hline
\hline
Model & Dataset & BLEU-1 & ROUGE-L\\
\hline
R-NET& MSMARCO & 41.29 & 43.38\\
R-NET with 2-level \textbf{Ham}& MSMARCO & 41.37 & 43.89\\
R-NET with 5-level \textbf{Ham}& MSMARCO & 43.13 & 45.67\\
R-NET with 10-level \textbf{Ham}& MSMARCO & 43.48 & 45.75\\
R-NET with 20-level \textbf{Ham}& MSMARCO & \textbf{43.62} & \textbf{45.78}\\
\hline
\end{tabular}
\caption{Evaluation results for MRC models}
\end{table}

\subsection{Chinese Poem Generation}
The results of our BLEU-based evaluation are summarized in Table 2. We compare our \textbf{Ham-based PPG} with several relevant baselines like \textbf{Statistical Machine Translation (SMT)} proposed by He, et al.\ (2012) and \textbf{RNN-based Poem Generator (RNNPG)} proposed by Zhang, et al.\ (2014). In the former model, a poem is generated iteratively by translating the previous line into the next line. In the latter model, all the lines are generated based on a context vector encoded from the previous lines. We also set different attention depths in order to learn the relationship between overall performance and the attention depth $d$.

\begin{table}[htb]
\centering
\begin{tabular}{|l|cc|cc|cc|cc|}
\hline
\multicolumn{1}{|l}{Model}&
\multicolumn{2}{|c|}{$\Bleu_{1}$}&
\multicolumn{2}{c|}{$\Bleu_{2}$}&
\multicolumn{2}{c|}{$\Bleu_{3}$}&
\multicolumn{2}{c|}{$\Bleu$}       \\
\cline{2-9}
\multicolumn{1}{|l}{} &
\multicolumn{1}{|c|}{5-Char}&{7-Char} &
\multicolumn{1}{c|}{5-Char}&{7-Char} &
\multicolumn{1}{c|}{5-Char}&{7-Char} &
\multicolumn{1}{c|}{5-Char}&{7-Char}    \\
\hline
\hline
SMT & 0.056 & 0.124 & 0.052 & 0.150 & 0.054 & 0.176 & 0.054 & 0.150 \\
RNNPG & 0.058 & 0.187 & 0.062 & 0.210 & 0.067 & 0.207 & 0.062 & 0.202 \\
PPG & 0.061 & 0.185 & 0.069 & 0.193 & 0.073 & 0.198 & 0.068 & 0.192 \\
\hline
5-level \textbf{Ham} PPG& \textbf{0.063} & 0.210 & 0.070 & 0.237 & 0.075 & 0.226 & 0.070 & 0.224\\
10-level \textbf{Ham} PPG& 0.062 & 0.217 & \textbf{0.075} & \textbf{0.267} & \textbf{0.076} & 0.259 & \textbf{0.072} & 0.244\\
20-level \textbf{Ham} PPG& 0.062 & \textbf{0.221} & 0.074 & 0.258 & \textbf{0.076} & \textbf{0.260} & 0.071 & \textbf{0.246}\\
\hline
\end{tabular}
\\
\caption{BLEU-based evaluation results}
\end{table}
%From the table, we learn that \textbf{Ham} does make some difference. It is  clear from the table that the proposed High-level \textbf{Ham-based PPG} outperforms conventional methods in all cases.
When attention depth $d$ is not large enough, the larger $d$ is, the better performance our model will achieve. \textbf{PPG} with 10-level \textbf{Ham} has almost $25\%$ of improvement on the averaged {BLEU} score compared with ordinary \textbf{PPG}. On the other hand, through the last two rows of the table above and Theorem 2, we know that with the continuing increase of $d$, our performance will converge sooner or later. So in order to balance between performance and training cost, we suggest attention depth $d$ to be between 5 and 10.
\begin{figure}[h]
\centering
\includegraphics[width=0.80\textwidth]{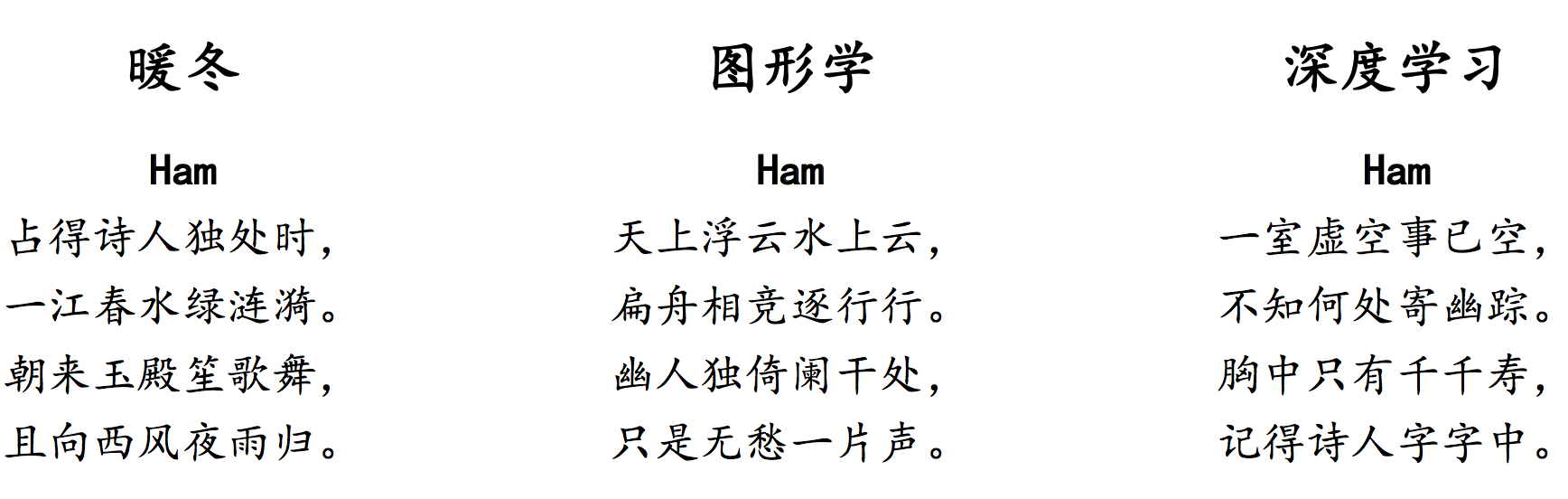}
\caption{Three quatrains generated by \textbf{10-level Ham PPG} model}
\end{figure}

\section{Concluding Remarks}

In this paper we have developed \textbf{Hierarchical Attention Mechanism (Ham)}. So far as we  know, This is the first attention mechanism which introduces hierarchical mechanisms into attention mechanisms and takes the weighted sum of different attention levels as the output, so it  combines low-level features and high-level features of input sequences to output a more suitable intermediate result for decoders.
%
%We tested the proposed model, called RNNsearch, on the task of English-to-French translation. The experiment revealed that the proposed RNNsearch outperforms the conventional encoder–decoder model (RNNencdec) significantly, regardless of the sentence length and that it is much more ro- bust to the length of a source sentence. From the qualitative analysis where we investigated the (soft-)alignment generated by the RNNsearch, we were able to conclude that the model can cor- rectly align each target word with the relevant words, or their annotations, in the source sentence as it generated a correct translation.

We tested the proposed model \textbf{Ham} on the task of Chinese poem generation and machine reading comprehension. The experiment revealed that the proposed  \textbf{Ham} outperforms the conventional models significantly, achieving the  state-of-the-art results. In the future, we would like to study more applications of \textbf{Ham} on other NLP tasks such as neural machine translation, abstractive summarization, paraphrase generalization and so on. 

Recall that \textbf{Ham} belongs to soft attention where every token of input sequences is calculated by attention function. 
 We will extend  \textbf{Ham} to hard attention and local attention, to show whether the performance can be better and whether \textbf{Ham} can fit reinforcement learning environment better. Also, we will attempt to extend \textbf{Multi-head Attention Mechanism}  of Vaswani, et al.\ (2017) to its hierarchical version and apply  
 to neural machine translation. 
 %Another feasible idea is the combination of \textbf{Ham} and ResNet by He, et al.\ (2016) which we will try later. Our goal is to make attention mechanism more powerful.

\section*{References}
\medskip

\small

[1] Vaswani, Ashish, Noam Shazeer, Niki Parmar, Jakob Uszkoreit, Llion Jones, Aidan N. Gomez, Lukasz Kaiser, and Illia Polosukhin. "Attention is all you need." In Advances in Neural Information Processing Systems, pp. 6000-6010. 2017.

[2] Bahdanau, Dzmitry, Kyunghyun Cho, and Yoshua Bengio. "Neural machine translation by jointly learning to align and translate." arXiv preprint arXiv:1409.0473 (2014).

[3] Shenjian Zhao and Zhihua Zhang. Attention-via-Attention Neural Machine Translation. In Proceedings of the Thirty-Second AAAI Conference on Artificial Intelligence (AAAI'18) , 2018.

[4] Lin, Zhouhan, Minwei Feng, Cicero Nogueira dos Santos, Mo Yu, Bing Xiang, Bowen Zhou, and Yoshua Bengio. "A structured self-attentive sentence embedding." arXiv preprint arXiv:1703.03130 (2017).

[5] Seo, Minjoon, Aniruddha Kembhavi, Ali Farhadi, and Hannaneh Hajishirzi. "Bidirectional attention flow for machine comprehension." arXiv preprint arXiv:1611.01603 (2016).

[6] Wang, Shuohang, and Jing Jiang. "Machine comprehension using match-lstm and answer pointer." arXiv preprint arXiv:1608.07905 (2016).

[7] Wang, Shuohang, Mo Yu, Xiaoxiao Guo, Zhiguo Wang, Tim Klinger, Wei Zhang, Shiyu Chang, Gerald Tesauro, Bowen Zhou, and Jing Jiang. "R $^ 3$: Reinforced Reader-Ranker for Open-Domain Question Answering." arXiv preprint arXiv:1709.00023 (2017).

[8] Wang, Wenhui, Nan Yang, Furu Wei, Baobao Chang, and Ming Zhou. "Gated self-matching networks for reading comprehension and question answering." In Proceedings of the 55th Annual Meeting of the Association for Computational Linguistics (Volume 1: Long Papers), vol. 1, pp. 189-198. 2017.

[9] Wang, Zhe, Wei He, Hua Wu, Haiyang Wu, Wei Li, Haifeng Wang, and Enhong Chen. "Chinese poetry generation with planning based neural network." arXiv preprint arXiv:1610.09889 (2016).

[10] Xu, Kelvin, Jimmy Ba, Ryan Kiros, Kyunghyun Cho, Aaron Courville, Ruslan Salakhudinov, Rich Zemel, and Yoshua Bengio. "Show, attend and tell: Neural image caption generation with visual attention." In International Conference on Machine Learning, pp. 2048-2057. 2015.

[11] Yiming Cui, Zhipeng Chen, Si Wei, Shijin Wang, Ting Liu and Guoping Hu. "Attention-over-Attention Neural Networks for Reading Comprehension" arXiv preprint arXiv:1607.04423v4 (2017).

[12] Rush, Alexander M., Sumit Chopra, and Jason Weston. "A neural attention model for abstractive sentence summarization." arXiv preprint arXiv:1509.00685 (2015).

[13] Mihalcea, Rada, and Paul Tarau. "Textrank: Bringing order into text." In Proceedings of the 2004 conference on empirical methods in natural language processing. 2004.

[14] Mikolov, Tomas, Kai Chen, Greg Corrado, and Jeffrey Dean. "Efficient estimation of word representations in vector space." arXiv preprint arXiv:1301.3781 (2013).

[15] He, Jing, Ming Zhou, and Long Jiang. "Generating Chinese Classical Poems with Statistical Machine Translation Models." In AAAI. 2012.

[16] Zhang, Xingxing, and Mirella Lapata. "Chinese poetry generation with recurrent neural networks." In Proceedings of the 2014 Conference on Empirical Methods in Natural Language Processing (EMNLP), pp. 670-680. 2014.

[17] He, Kaiming, Xiangyu Zhang, Shaoqing Ren, and Jian Sun. "Deep residual learning for image recognition." In Proceedings of the IEEE conference on computer vision and pattern recognition, pp. 770-778. 2016.

[18] He, Wei, Kai Liu, Yajuan Lyu, Shiqi Zhao, Xinyan Xiao, Yuan Liu, Yizhong Wang et al. "DuReader: a Chinese Machine Reading Comprehension Dataset from Real-world Applications." arXiv preprint arXiv:1711.05073 (2017).

[19] Yang, Zichao, Diyi Yang, Chris Dyer, Xiaodong He, Alex Smola, and Eduard Hovy. "Hierarchical attention networks for document classification." In Proceedings of the 2016 Conference of the North American Chapter of the Association for Computational Linguistics: Human Language Technologies, pp. 1480-1489. 2016.

\end{document}